\documentclass[10pt,journal,onecolumn]{IEEEtran}
\usepackage{graphicx}
\usepackage{epsfig}
\usepackage{epstopdf}
\usepackage{float}
\usepackage{algorithmic}
\usepackage{algorithm}
\usepackage{bm}


\begin{document}
%
\title{ODMTCNet: An Interpretable Multi-view Deep Neural Network Architecture for Image Feature Representation}
%
%
%

\author{Lei~Gao,~\IEEEmembership{Member,~IEEE,}
        Zheng~Guo,
        and~Ling~Guan,~\IEEEmembership{Fellow,~IEEE}
\thanks{L. Gao, Z. Guo and L. Guan are with the Department of Electrical and Computer Engineering, Ryerson University, Toronto, ON M5B 2K3, Canada (email:iegaolei@gmail.com; zjguo@ryerson.ca; lguan@ee.ryerson.ca).}}

\maketitle

\begin{abstract}
Recently, deep cascade architecture-based algorithms have attracted wide interest and have been applied to various application domains successfully. However, the longstanding challenge of interpretability, is still considered as an Achilles' heel of such algorithms. Moreover, due to its data-driven nature, the deep cascade architecture likely causes over-fitting problems when there is no sufficient data available. To address these pressing issues, this work proposes an interpretable multi-view deep neural network architecture, namely optimal discriminant multi-view tensor convolutional network (ODMTCNet), by integrating statistical machine learning (SML) principles with the deep neural network (DNN) architecture. Benefiting from the joint strength of SML and DNN, we demonstrate that ODMTCNet is analytically interpretable for multi-view image feature representation. Specifically, a discriminant multi-view tensor convolution strategy is proposed and integrated with the desired deep cascade architecture to generate high quality feature representations. Different from the traditional DNN models, the parameters of the convolutional layers in ODMTCNet are determined by analytically solving a SML--based optimization problem in each convolutional layer independently. This work demonstrates that, in ODMTCNet, the relation between the optimal performance and parameters (e.g., the number of convolutional filters) can be predicted, with each layer generating justified knowledge representations, leading to an interpretable multi-view based convolutional network. In addition, an information theoretic based descriptor, information quality (IQ), is utilized for feature representation of the given multi-view data sets. Because of its unique design, ODMTCNet is able to handle image data sets of different scales, large or small, effectively addressing the data hungry nature of DNN in image representation and forming a generic platform for multi-view image feature representation. To validate the effectiveness and the generic nature of the proposed ODMTCNet, we conducted experiments on four image data sets of different scales: The Olivetti Research Lab (ORL) database, Facial Recognition Technology (FERET) database, ETH--80 database and Caltech 256 database. The results show the superiority of the proposed solution compared to state-of-the-art.
\end{abstract}


%
\IEEEpeerreviewmaketitle
\section{Introduction}
With the advancement of large scale data-based techniques (e.g., big data, deep learning, etc.), there has come urgent demand to better understand and reveal the intrinsic properties of data structure  in order to effectively explore useful information from massive sets of data. As a solution to this challenge, feature representation has been attracting widespread attention and investigation [1-3]. In general, the main streams of feature representation are divided into three different categories: handcrafted feature representation, prior knowledge-based feature representation and learned deep-level feature representation. In what follows, the three different categories of feature representation are introduced briefly.\\\indent Due to its ability to extract geometry and texture characteristics of given samples reliably (such as local binary pattern, scale-invariant feature transforms and gradient orientation histogram, etc.), handcrafted feature representation methodology has drawn tremendous interest and has been utilized in a variety of real applications. However, suffering from the intra-class modifiability problem [4], handcrafted methodology is difficult to perform well on unseen samples. To tackle this problem, prior knowledge-based feature representation solutions have been proposed and investigated extensively, such as manifold representation [5], subspace representation [6] and dictionary representation [7], etc. Generally, according to certain prior knowledge, the studied distinct information from given samples can be applied to new or unseen samples successfully, leading to better performance. It should be noted that, although more efforts have been committed to the development of prior knowledge-based feature representation algorithms, plenty of room is still available for further improvement, especially when working with large scale data sets [84-86].\\\indent In recent years, by generating learned deep-level feature representation, deep neural network (DNN)-based algorithms have opened up a new frontier to promote state-of-the-art in visual computing, image classification and many other computer vision tasks [8-10]. Compared with the handcrafted and prior knowledge-based feature representation strategies, deep-level feature representation attempts to mimic the workings of the human brain in processing data/information. In general, deep-level feature representation is generated by utilizing a multi-layer cascade architecture and the learning process needs enormous data samples to guarantee that the parameters in the deep layers are tuned properly. Particularly, there are two outstanding issues, among others, with the current deep learning architectures: (a) Model interpretability, a persistent challenge since the inception of DNN research; (b) Over-fitting in the learned deep-level feature representation when there is no sufficient data available due to DNNs` data-driven nature.\\\indent One recently proposed solution to model interpretability is to modify a DNN model so that each convolutional layer generates knowledge representations, such as the loss method presented in [91] and an interpretable feedforward (FF) solution in [92]. In addition, the integration of prior knowledge with DNN-based architectures is becoming an increasingly popular research area for tackling the model interpretability problem (e.g., PCANet [11], DCTNet [12], etc.). Amongst these solutions, the model parameters are usually calculated by a certain prior knowledge-based function while the final outputs are generated by the deep cascade architecture.\\\indent On the other hand, to address the over-fitting problems in DNN-based models, the multi-view feature representation solutions have attracted enormous attention. [13] explores the integration of multi-view features for better data augmentation to tackle small sample size (SSS) problems, and yields improved performance. The study in [13] suggests that one desirable solution to balancing the small scale and deep-level feature representation is to smartly utilize multi-view features and the deep cascade structure. A canonical correlation analysis network (CCANet) is presented in [4]. CCANet can obtain more comprehensive information from the inputs, resulting in enhanced performance. However, due to its unsupervised nature, CCANet lacks the power to extract discriminant information of high quality, leading to suboptimal performance. To address this problem, a discriminative canonical correlation network (DCCNet) [14] is introduced. However, in the DCCNet, only between-class and within-class scatter matrices are employed. Since the purpose of the scatter matrices is simply to reveal and extract discriminant representations of the given samples from a single data set, such a matrix is incapable of gaining the discriminant information across different sets of features adequately. Therefore, DCCNet essentially cannot be applied to multi-view feature representation directly.\\\indent In this paper, an interpretable multi-view deep learning architecture, namely optimal discriminant multi-view tensor convolutional network (ODMTCNet), is proposed to address the model interpretability and over-fitting problems jointly. ODMTCNet is essentially based on the integration of statistical machine learning (SML) principles with DNN architecture. Specifically, in ODMTCNet, a discriminant multi-view tensor convolutional strategy is proposed and applied to the multi-view data sources. In addition, correlation matrices instead of scatter matrices are utilized and optimized to explore the discriminant information across different data sets. Note that, unlike the scatter matrices, the power of which is to reveal the discriminant relation among various samples in a single data set, the correlation matrices are capable of exploring the discriminant information across multiple data sets. Thus, the correlation matrices can more adequately explore the discriminant and intrinsic relation amongst different data sets in multi-view feature representation. In other words, the proposed ODMTCNet potentially generates a high quality feature representation for image based applications. The difference between scatter matrices and correlation matrices is depicted graphically in Figure. 1.\\\indent Since there is no loss function with given/fixed values associated with ODMTCNet, the popular backpropagation (BP) algorithm is not used in this work and the parameters of the convolutional layers in the network are determined by analytically solving a SML-based optimization problem in each convolutional layer independently instead. Moreover, according to SML principles, the relation between the optimal performance and parameters (e.g., the number of convolutional filters in each layer) in ODMTCNet is predicted analytically, providing a simple and appealing alternative to the popular try-and-error practice in DNN architecture design. Hence, each convolutional layer in the ODMTCNet model generates justified knowledge representations, making the proposed model interpretable.\\\indent In addition, by utilizing the designed discriminant multi-view tensor convolution, ODMTCNet yields more distinct multi-view feature maps for data augmentation, providing a solution for improving performance on small scale data sets. Due to the characteristics of the deep cascade architecture, ODMTCNet can also exploit abstract semantics from the large scale data sources, and therefore works well on large-scale sets as well. Finally, an information theoretic based descriptor, information quality (IQ) is introduced to generate the high quality feature representation from the original multi-view data sources. To validate the effectiveness and generality of the proposed network, we conduct experiments on four image data sets of different scales: the Olivetti Research Lab (ORL) database, Facial Recognition Technology (FERET) database, ETH--80 database and Caltech 256 database. Experimental results show the superiority of ODMTCNet over state-of-the-art for multi-view feature representation.\\\indent This paper is organized as follows: A review of related work is given in Section 2. The proposed ODMTCNet is presented in Section 3. Experiments, comparison with state-of-the-art and discussions are conducted in Section 4. Section 5 draws the conclusions.
\begin{figure*}[t]
\centering
\includegraphics[height=1.5in,width=7.25in]{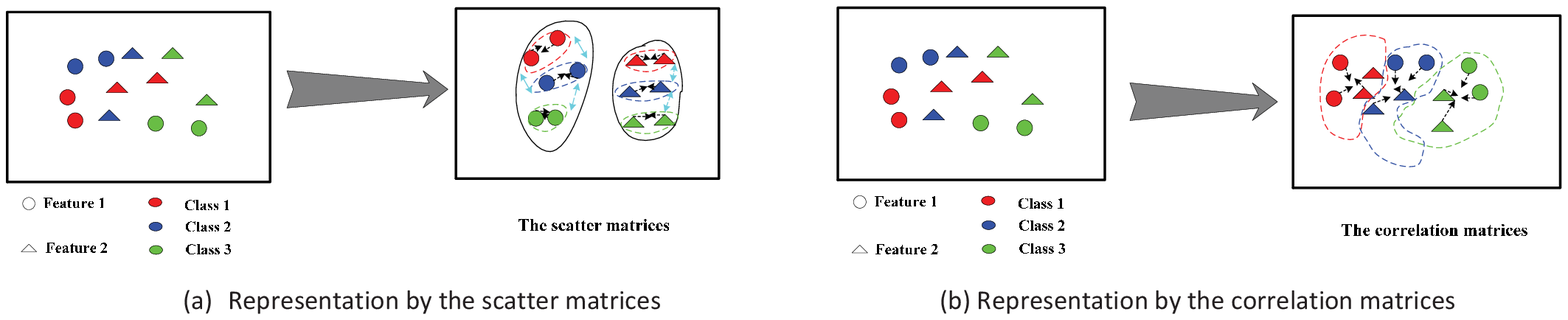}\\ Figure. 1 The representation by scatter matrices and correlation matrices.\\
\end{figure*}
\section{Related Work}
In this section, we briefly present related work on convolutional neural networks (CNN), traditional correlation analysis (TCA) and discriminant correlation analysis (DCA) methods, laying theoretical foundations for the proposed ODMTCNet.
\subsection{CNN}
Motivated by biological processes of the human visual cortex, CNN is proposed and applied to visual computing, video processing, image classification, among others. A typical CNN architecture usually includes convolution layers, pooling layers and fully connected layers as drawn in Figure. 2.
\begin{figure}[H]
\centering
\includegraphics[height=1.4in,width=3.4in]{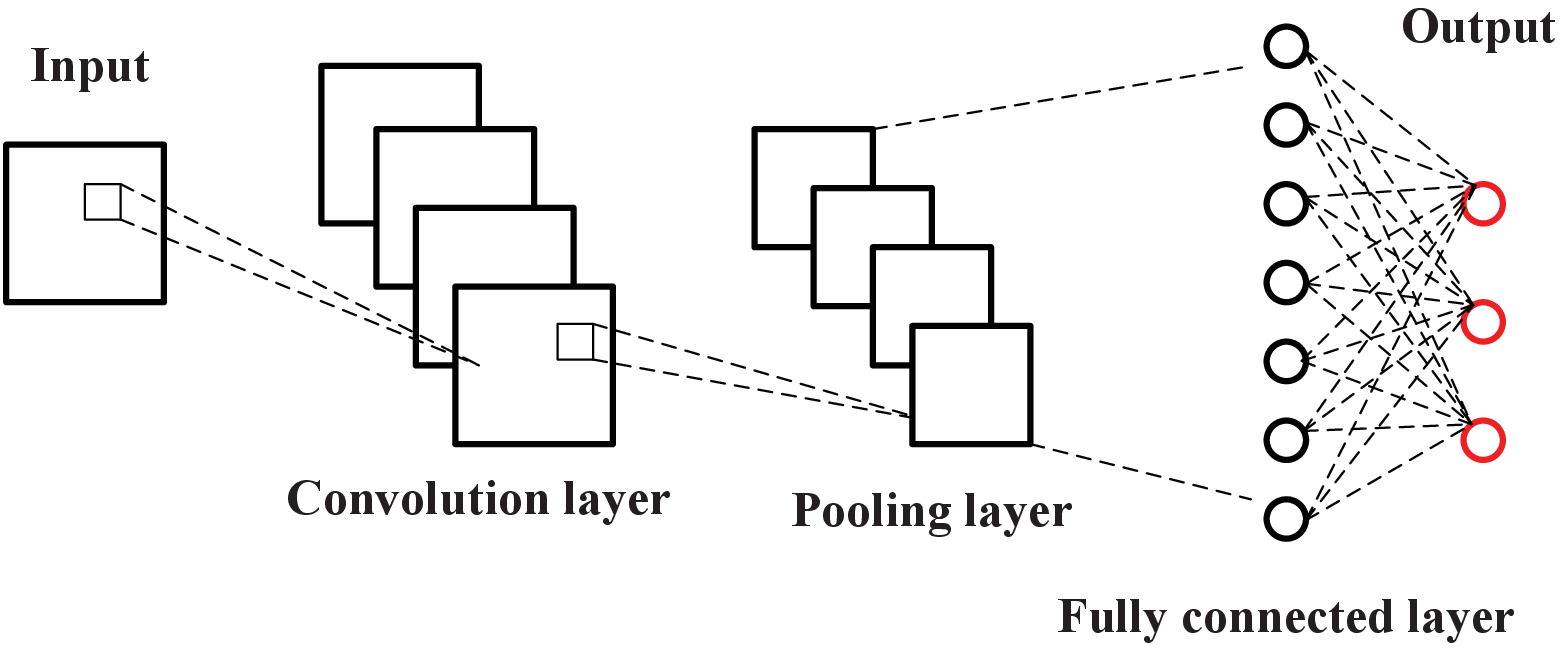}\\ Figure. 2 A typical CNN architecture
\end{figure}
Generally, the feedforward and backpropagation algorithms are utilized in CNN to adjust the internal parameters to reduce the targeted loss function, resulting in better performance. As the number of layers increases, enormous data samples are required to guarantee that the parameters in different layers are tuned properly for the intended tasks. Thus, a CNN is unlikely to produce satisfactory results with small scale inputs.
\subsection{TCA}
TCA aims to find projected vectors for the given data sets to obtain the maximum correlation. Assume there are two data sets $x$ and $y$. Then, the objective of TCA is to seek a pair of projected vectors $ \omega_{x} $ and $ \omega_{y} $ with the maximum correlation. Or mathematically, TCA solves the following optimization problem
\begin{equation} \mathop {\arg \max }\limits_{{\omega _{x}},{\omega _{y}}} {\omega _{x}}^T{R_{xy}} {\omega _{y}},\end{equation}
\emph{s.t}
\begin{equation} {\omega _{x}}^T{R_{xx}}{\omega _{x}} = {\omega _{y}}^T{R_{yy}}{\omega _{y}} = 1, \end{equation}
where ${R_{xy}} = x{y^T}$, ${R_{xx}} = x{x^T}$ and ${R_{yy}} = y{y^T}$. The solutions to equations (1) and (2) are gained by the eigenvalue decomposition or singular value decomposition. Nevertheless, due to the unsupervised nature, it usually leads to an unsatisfying performance.
\subsection{DCA}
Assume there are two sets of random variables ${x_1}^{\sim} \in {R^{{m_1} \times N}},{x_2}^{\sim} \in {R^{{m_2} \times N}}$ as the entries with $N$ being the total number of training samples, and ${m_1}$ \& ${m_2}$ are the respective dimensionalities of the two sets. Then, the zero-mean sets of ${x_1}^{\sim}$ and ${x_2}^{\sim}$ are written as $ {x_1} = {x_1}^{\sim} - {M_{x_1}} $ and ${x_2} = {x_2}^{\sim} - {M_{x_2}}$ with ${M_{x_1}}$ and ${M_{x_2}}$ being the mean matrices in equation (3)
\begin{equation}
{M_{x_1}} = \frac{1}{N}\sum\limits_{i = 1}^N {{x_{1i}}^{\sim}}, {M_{x_2}} = \frac{1}{N}\sum\limits_{i = 1}^N {{x_{2i}}^{\sim}},
\end{equation}
where ${x_{1i}}^{\sim}$ and ${x_{2i}}^{\sim}$ are the $i$th sample in ${x_{1}}^{\sim}$ and ${x_{2}}^{\sim}$, respectively. The within-class and between-class correlation matrices of two sets $x_1$ and $x_2$ are expressed as ${C_{{w_{{x_1}{x_2}}}}}$ and ${C_{{b_{{x_1}{x_2}}}}}$, which are formulated as ${C_{{w_{{x_1}{x_2}}}} = {x_1}D{x_2}^T}$ and ${{C_{{b_{{x_1}{x_2}}}}}} =  - {x_1}D{x_2}^T$ [15], where
\begin{equation} \ D = \left[ {\left( {\begin{array}{*{20}{c}}{{H_{{n_{i1}} \times {n_{i1}}}}}& \ldots &0\\
 \vdots &{{H_{{n_{il}} \times {n_{il}}}}}& \vdots \\
0& \ldots &{{H_{_{{n_{ic}} \times {n_{ic}}}}}}
\end{array}} \right)} \right] \in {R^{N \times N}}, \end{equation}
and ${n_{il}}(l \in [1,2,...c])$ is the number of samples in the \textit{l}th class within the $i$th data set $x_i$ and ${H_{{n_{i{l}}} \times {n_{il}}}}$ is a square matrix of size ${n_{il}} \times {n_{il}}$ with unit values for all the elements.\\\indent DCA aims to find a pair of
projected vectors ${w_1}$ and ${w_2}$ with the discriminant representations by utilizing the within-class and between-class correlation matrices across two data sets, and its optimization problem is formulated as below:
\begin{equation} \mathop {\arg \max \lambda }\limits_{{w_1},{w_2}}  = {{w_1}^T\mathop {{C_{{x_1}{x_2}}}}\limits^ \sim  {w_2}} {\rm{}}, \end{equation}
subject to
\begin{equation} {{w_1}^T{C_{{x_1}{x_1}}}{w_1}}  = {{w _2}^T{C_{{x_2}{x_2}}}{w _2}} = 1, \end{equation}
where $\mathop {{C_{{x_1}{x_2}}}} \limits^ \sim  = {{C_{{w_{{x_1}{x_2}}}}}}-{{C_{{b_{{x_1}{x_2}}}}}}$, ${\rm{ }}{C_{{x_1}{x_1}}} = {x_1}{x_1}^T$ and ${\rm{ }}{C_{{x_2}{x_2}}} = {x_2}{x_2}^T$. Then the Lagrange multiplier method and eigenvalue decomposition (GEV) algorithm are utilized to find the solution to equation (5). Essentially, DCA is a special case of discriminative multiple canonical correlation analysis (DMCCA) when only two data sets are utilized [15]. Moreover, there exists an important property of DCA that the number of projected dimensions \emph{d} associated with the global optimal performance is smaller than or equals to the number of classes, \emph{c}, or mathematically [15]
 \begin{equation} \ d \le c .\end{equation}
\section{The proposed ODMTCNet}
The proposed ODMTCNet consists of three types of layers: DCA driven multi-view convolutional layers, hashing controlled pooling layers and fully connected layers with information quality enhancement. Essentially, the ultimate goal of ODMTCNet is to generate a novel multi-view feature representation on image data sets of different scales. In what follows, the construction and design of these layers are presented, consecutively. Afterwards, we move on to the analysis of relationship between parameters (e.g., the number of convolutional filters) and the optimal performance.

\subsection{The DCA driven multi-view convolutional layer}
The convolutional layer plays a vital role in the whole CNN architecture, which is accomplished by conducting the convolution operation between inputs and the pre-defined convolutional filters [16].\\\indent Given a set of images ${I = [{I_{1}},{I_{2}},...{I_{M}}]}$, where $M$ is the number of images, assume there are two views of this data set, ${I^{1,M}=[{I^{1}_{1}},{I^{1}_{2}},...{I^{1}_{M}}]}$ and ${I^{2,M}=[{I^{2}_{1}},{I^{2}_{2}},...{I^{2}_{M}}]}$. Then, the sample pair from ${I^{1,M}}$ and ${I^{2,M}}$ is a pair of images with a size of $p \times q$ pixels as shown in Figure. 3.
\begin{figure}[H]
\centering
\includegraphics[height=0.9in,width=2.8in]{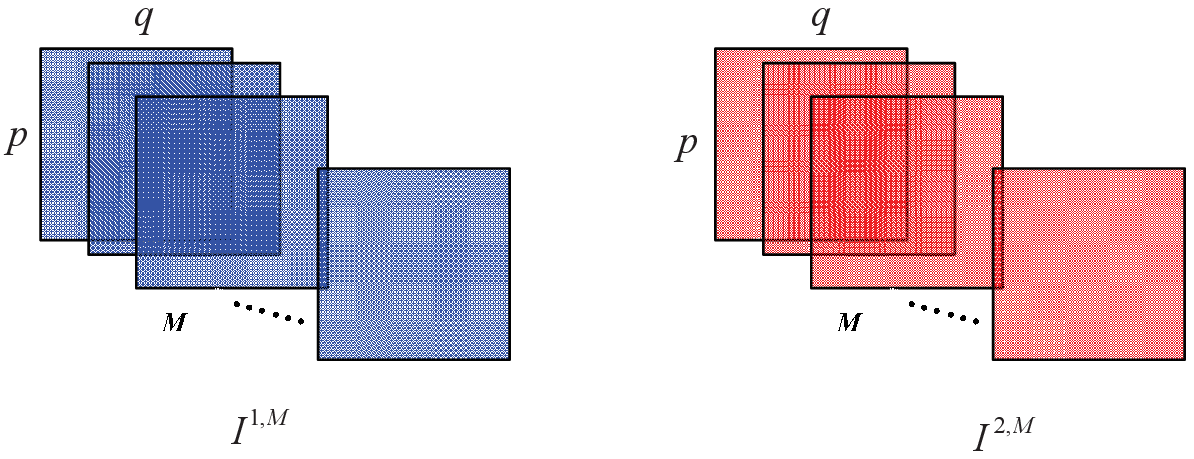}\\ Figure. 3 Representation of two-view data sets
\end{figure}
In Figure. 3, it is observed that $I^{1,M}$ and $I^{2,M}$ can be presented by two three-dimensional tensors. In the following, our target is to explore the discriminant feature representation between the two three-dimensional tensors. First, each sample pair in the given two-view data sets is divided into patches with the size of ${l_1} \times {l_2}$. Next, we select an ${l_1} \times {l_2}$ patch around each pixel and the vectorized operation is performed on the selected patches, yielding a vector of length ${l_1}{l_2}$. Thus, all patches collected from the $k$th sample of the $d$th ($d \in [1,2]$) view set are written in the vectorial forms ${{I^{d}_{k,1}},{I^{d}_{k,2}},...,{I^{d}_{k,pq}}} \in {R^{{l_1} {l_2}}}$. The patches from all $M$ samples of the $d$th view set are expressed as follows
\begin{equation}
I^{d} = [{I^{d}_1}^*,{I^{d}_2}^*,...,{I^{d}_M}^*] \in {R^{{l_1}{l_2} \times Mpq}},
\end{equation}
where
\begin{equation}
{I^{d}_k}^* = [{I^{d}_{k,1}},{I^{d}_{k,2}},...,{I^{d}_{k,pq}}] \in {R^{{l_1}{l_2} \times pq}} (k \in [1,2,...,M]).
\end{equation}
In ODMTCNet, the optimization function of the DCA method is to learn the parameters of the cascade network from the input samples to explore discriminant feature representation. $I^{d}$ is considered as the $d$th data set with size ${l_1}{l_2} \times Mpq$, where ${l_1}{l_2}$ is the number of `dimensions' and $Mpq$ is the number of `training samples', respectively. As a result, the optimization function of DCA on $I^{1}$ and $I^{2}$ is formulated as follows
\begin{equation} \mathop {\arg \max \rho }\limits_{{\omega _1},{\omega _2}}  = {{\omega _1}^T\mathop {{C_{{I^{1}}{I^{2}}}}}\limits^ \sim  {\omega _2}} {\rm{}}, \end{equation}
subject to
\begin{equation} {{\omega _1}^T{C_{I^{1}I^{1}}}{\omega _1}}  = {{\omega _2}^T{C_{I^{2}I^{2}}}{\omega _2}} = \textup{I}. \end{equation}
where $\omega _1$ \& $\omega _2$ are projected matrices, and $\textup{I}$ is an identity matrix. Based on the Lagrange multiplier method and the GEV algorithm, the solutions to $\omega _1$ and $\omega _2$ are of size ${l_1}{l_2} \times {l_1}{l_2}$. Then, the corresponding column pair in $\omega _1$ and $\omega _2$ is chosen as one convolutional filter for the ODMTCNet model, associated with the first view and second view respectively. As a result, ${l_1}{l_2}$ convolutional filters are generated according to different columns in $\omega _1$ and $\omega _2$. The utilization of these convolutional filters will yield ${l_1}{l_2}$ distinct feature maps, generating rich data augmentation which in turn enables the proposed method to improve the performance on small scale data sets. The design and construction of the convolutional filters are presented below. \\\indent Suppose we have $L_i$ filters in the $i$th layer of the $1$th view, the convolutional filters are formulated as below
\begin{equation}
W_g^{1} = re{s_{{l_1},{l_2}}}({{\omega _{1, g}}}) \in {R^{{l_1} \times {l_2}}},g = 1,2,..,{L_i} ({L_i}<{l_1}{l_2}),
\end{equation}
where $re{s_{{l_1},{l_2}}}(h)$ is a function to reshape $h \in {R^{{l_1}{l_2}}}$ into a matrix with size ${{l_1}\times {l_2}}$ and $\omega _{1, g}$ is the $g$th column in $\omega _1$. Similarly, the convolutional filters in the $2$th view are formulated as
\begin{equation}
W_g^{2} = re{s_{{l_1},{l_2}}}({{\omega _{2, g}}}) \in {R^{{l_1} \times {l_2}}},g = 1,2,..,{L_i}.
\end{equation}
Therefore, in the $i$th layer of the $d$th view, we obtain $L_i$ feature maps $I^{out}_{d, k, g}=re{s_{{l_1},{l_2}}}({I^{d}_k}^*) \otimes W_g^{d}$ from the $k$th sample, where `$\otimes$' is the 2D convolution operator. The aforementioned procedure is clearly visualized in Figure. 4.\\
\centerline {\includegraphics[height=0.8in,width=3.6in]{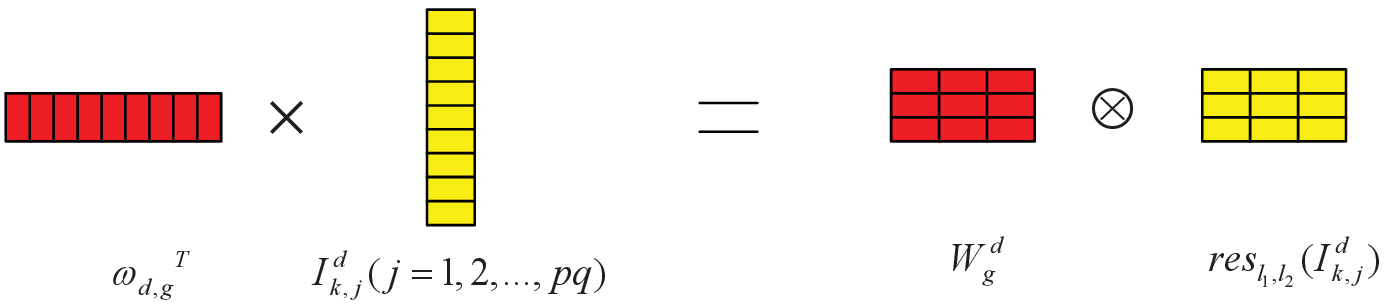}}\\ \centerline {Figure 4. 2D convolution and vectors-based product operation}\\\indent

The above analysis shows that the feature maps from the 2D convolutional filters are generated by solving a SML-based optimization function of DCA. Since the SML-based DCA is based on mathematically/statistically solid foundation with clearly defined optimization objectives, each convolutional layer in the ODMTCNet yields and possesses justified knowledge representations. By repeating this procedure, a DNN-based cascade network is built to generate the deep-level feature representation from the two-view data sets.
\subsection{The pooling layer built on multi-view hashing}
Based on the previous subsection, suppose there are $L_i$ convolutional filters in the $i$th layer from each view and each of them corresponds to the $L_{i+1}$ filters in the $i+1$th layer. Then, there are $L_i \times L_{i+1}$ feature maps in the $i+1$th layer from each view. Since a cascaded network structure is employed in ODMTCNet, the total number of feature maps from convolutional filters will increase greatly with the growing of the deep network layers as depicted graphically in Figure. 5. In the figure it can be seen that, with the increase of convolutional filters and layers, the proposed ODMTCNet potentially causes a slow convergence or overfilling problem. To address this problem, a new multi-view based hashing pooling layer is utilized in this subsection.\\
\centerline {\includegraphics[height=1.2in,width=2.8in]{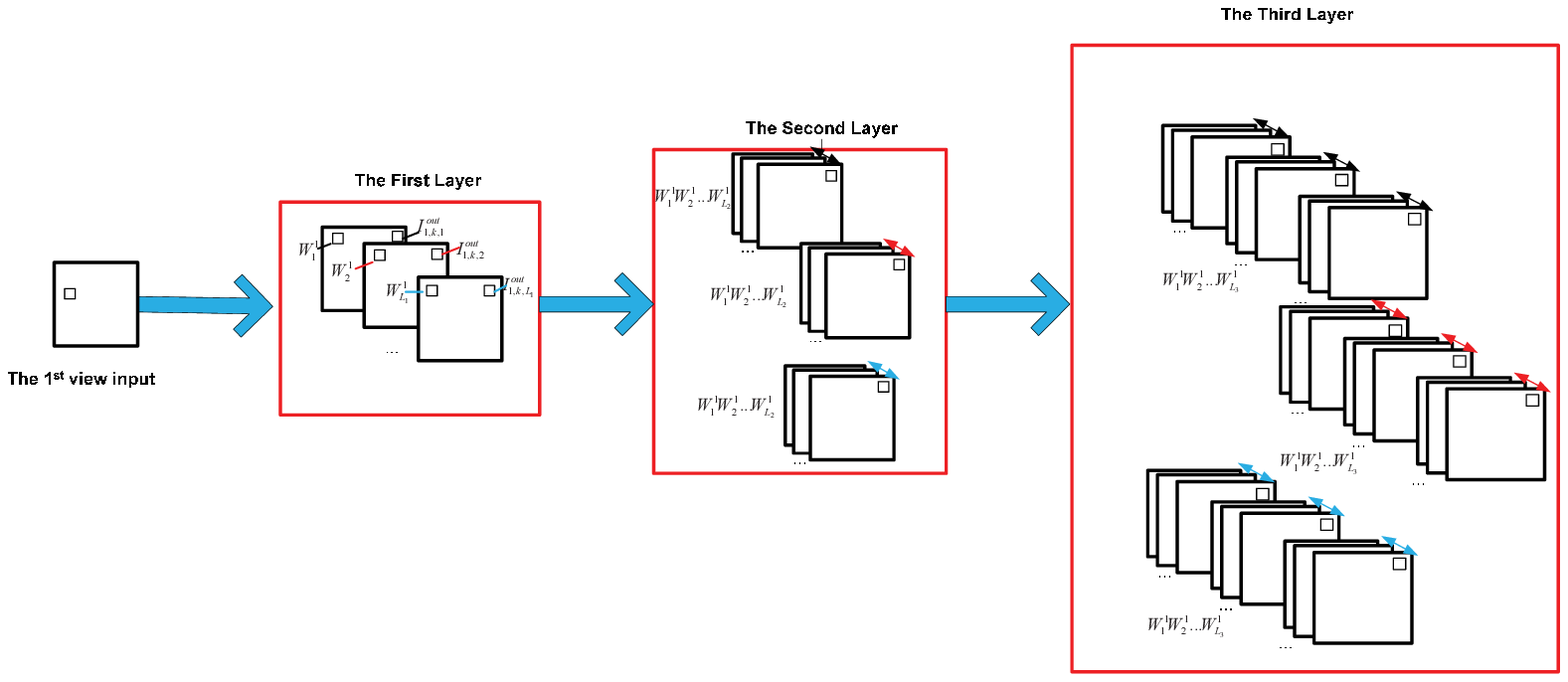}}\\ \centerline {Figure 5. Relation between feature maps and layers}\\
\begin{figure*}[t]
\centering
\includegraphics[height=2.0in,width=7.2in]{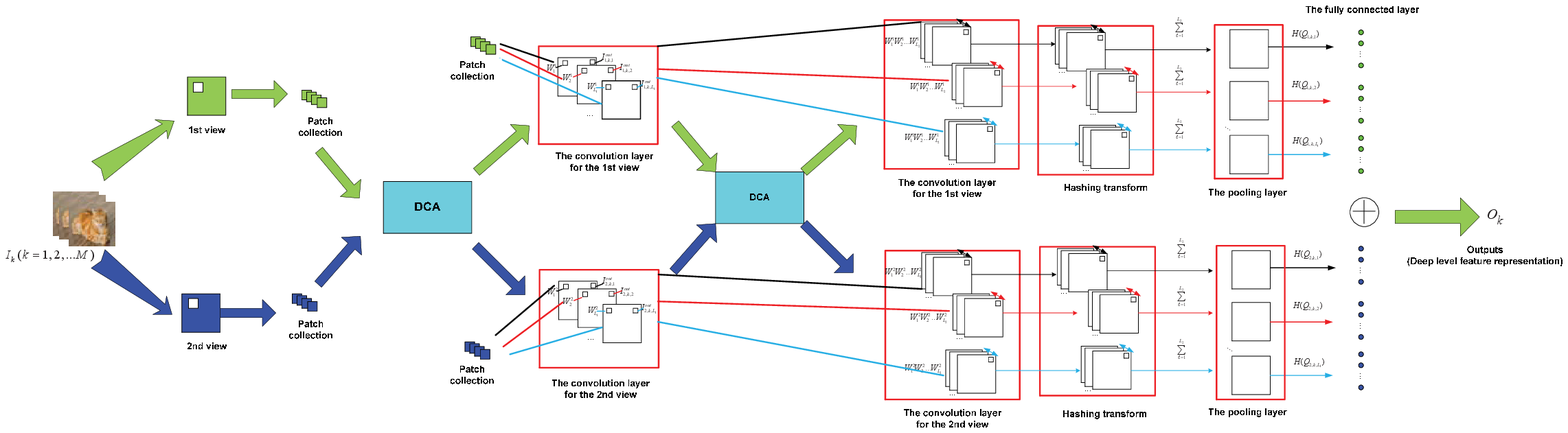}\\ Figure. 7 The overall architecture of ODMTCNet.\\
\end{figure*}\\\indent
Essentially, the pooling layer is a specific block of a typical CNN architecture. It aims to reduce the size of the feature representation from the previous layers, leading to smaller parameters and less computation in the related networks. The popular methods include max pooling, average pooling, K-max pooling, etc. In this paper, the pooling strategy is described as follows [11].\\\indent Assume there are $i+1$ layers in the proposed ODMTCNet architecture, and the outputs are expressed as $I^{out}_{d, k, g}  \otimes W_{\ell}^{d}$ ($\ell  = 1,2,..,{L_{i + 1}}$) for the $k$th sample of the $d$th view, where $I^{out}_{d, k, g}$ are the outputs of the $i$th layer ($g=1,2,..,{L_i}$). Then outputs are binarized based on Hashing transform in (14)
\begin{equation}
S\{ I^{out}_{d, k, g}  \otimes W_{\ell}^{d}\} (\ell  = 1,2,..,{L_{i + 1}}),
\end{equation}
where
\begin{equation}
S(x) = \left\{ \begin{array}{l}
 1,\quad (x > 0) \\
 0,\quad others. \\
 \end{array} \right.
\end{equation}
In essence, equation (15) can be considered as the activation function in a traditional CNN architecture. Next, the vector of $\ell$ binary bits is considered as a decimal number, resulting in a single integer-valued \emph{``image"} in (16)
\begin{equation}
Q_{d, k, g, \ell} = {{2^{\ell  - 1}}S\{ I^{out}_{d, k, g} \otimes W_{\ell}^{d}\} }.
\end{equation}
However, since there are $L_{i + 1}$ filters corresponding to each output from the previous layer, a pooling operator is in high demand to reduce the computational complexity efficiently. In this paper, equation (17) is adopted to accomplish the required `pooling' operation
\begin{equation}
Q_{d, k, g} = \sum\limits_{\ell  = 1}^{{L_{i + 1}}} Q_{d, k, g, \ell} = \sum\limits_{\ell  = 1}^{{L_{i + 1}}} {{2^{\ell  - 1}}S\{ I^{out}_{d, k, g} \otimes W_{\ell}^{d}\} },
\end{equation}
with each pixel being an integer in the range [$0, {2^{{L_{i + 1}}}} - 1$]. As a result, all the $L_{i + 1}$ images are combined into a single image, reducing the number of parameters and computation greatly. This procedure is depicted graphically in Figure. 6.\\ \hspace*{\fill} \\
\centerline {\includegraphics[height=1.6in,width=2.3in]{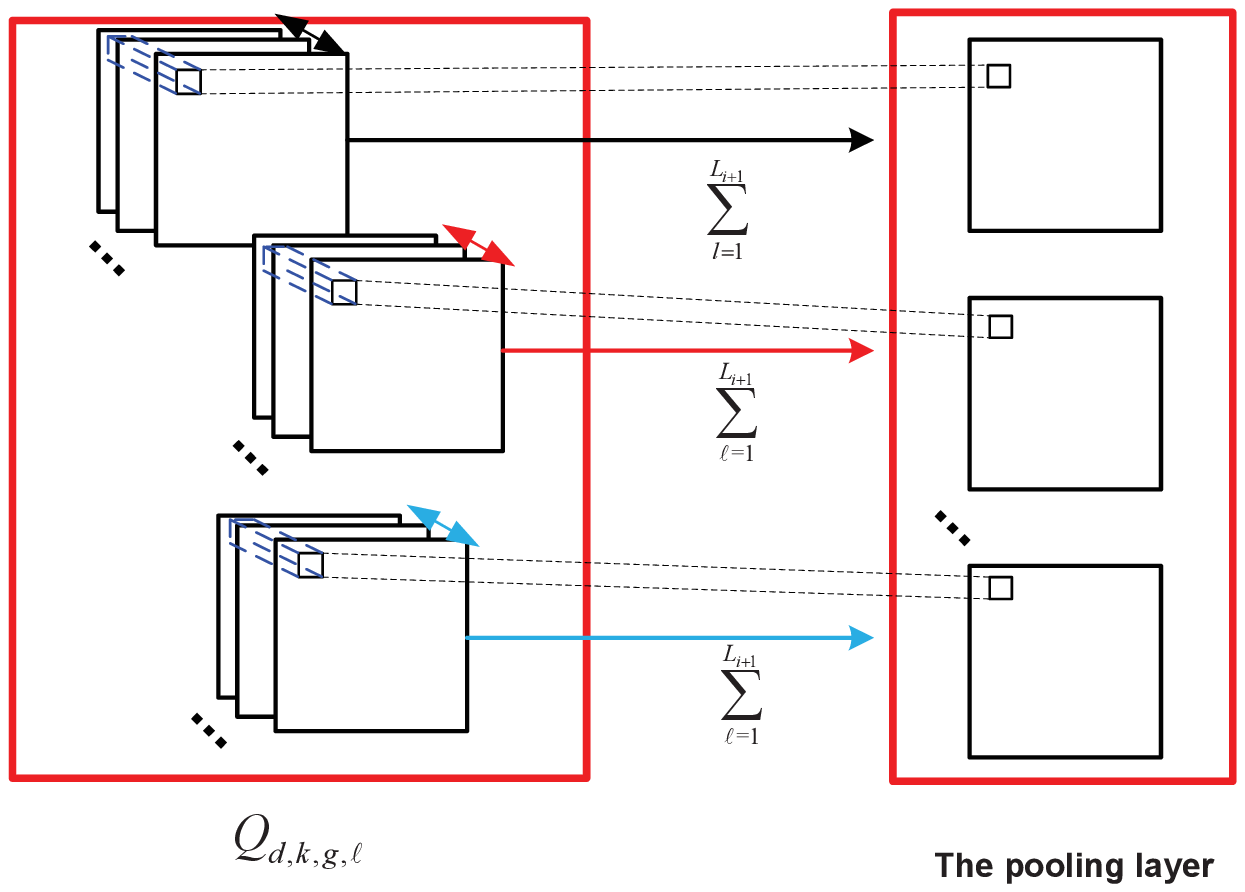}}\\ \centerline {Figure 6. The pooling layer}
\subsection{The information quality enhanced fully connected layer}
The fully connected layer is utilized to connect every neuron in one layer to every neuron in another layer. In this paper, an information quality (IQ)--based fully connected layer is proposed as follows to accomplish this task.\\\indent Based on the previous subsection, each generated image $Q_{d, k, g}$ is partitioned into $A$ blocks. Different from the existing CCANet and related algorithms, information quality (IQ) instead of histogram is employed to generate the feature representation. As a non-linear descriptor, IQ has the potential to represent the intrinsic structure of input data/information more powerfully. The definition of IQ is given in (18)
\begin{equation}
H(p(t)) = -log(p(t)),
\end{equation}
where $p(t)$ is the prior probability of $t$. 

Then, the IQ of the decimal values in each block is calculated and IQ values in all the $A$ blocks with an overlapping ratio ($\eta$) are transformed into one vector as $H(Q_{d, k, g})$. In ODMTCNet, the representation of the $k$th sample in the $d$th view is written in equation (19)
\begin{equation}
{o_{k, d}} = {[H(Q_{d, k, 1}),...,H(Q_{d, k, {L_i}})]^T} \in {R^{({2^{{L_{(i + 1)}}}}{L_i}A)}}.
\end{equation}
The fully connected feature representation corresponding to the $k$th sample by integrating two different views is formulated in (20)
\begin{equation}
{o_{k}} = [{o_{k, 1}}; {o_{k, 2}}] \in {R^{({2^{L_{(i+1)}+1}}{L_i}A)}}.
\end{equation}
The proposed feature representation architecture is illustrated in Figure. 7. This figure demonstrates that due to the combined strength of SML and DNN, not only are more discriminant feature maps generated, but abstract and robust semantics from large scale data sources are explored, yielding a high quality feature representation. Thus, by integrating data augmentation with more abstract and robust semantics, the proposed network works well on both small scale and large scale data sets.
\subsection{Relation between the optimal performance and parameters}
Having presented the basics in the proposed ODMTCNet, the relationship between the optimal performance and parameters (e.g., the number of convolutional filters in each layer) is presented below. According to the property of DCA in equation (7), the relation between the upper limit on the number of projected dimensions and the best solution is revealed as $\ d \le c$, where $d$ is the projected dimension associated with the global optimal performance and $c$ is the number of classes. Based on the mathematical analysis and Figure 4, it is known that the output from convolution filters is equal to the projection in DCA space as shown in Figure 8.\\
\centerline {\includegraphics[height=1.6in,width=3.6in]{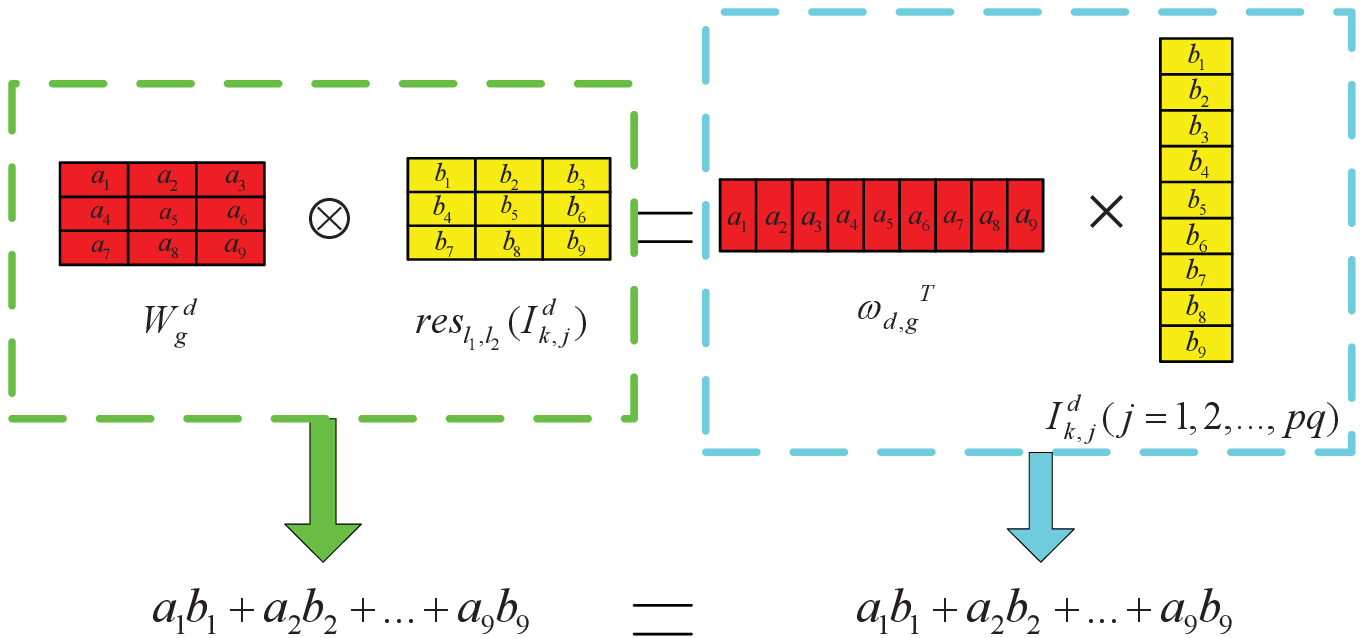}}\\ \centerline {Figure 8. The convolutional output and projection in DCA}\\

Since Figure 8 verifies that the output from convolution filters equals to the projection in DCA space, the outputs (e.g., feature maps) from the 2D convolutional filters also satisfy the property in equation (7). Therefore, the number of convolutional filters in each layer of ODMTCNet and the optimal performance can be predicted, leading to a discriminant representation. This is clearly a desirable and powerful feature of interpretable deep cascade networks.\\\indent The above analysis validates that since each convolutional layer in the ODMTCNet model generates discriminant knowledge representations, this work yields an interpretable DNN-based model on multi-view image feature representation. Note, as the discriminative representations are extracted according to an optimization function instead of a given/fixed loss function in each layer independently, the traditional BP algorithm is not required in the training process. By utilizing the designed discriminant multi-view tensor convolutional strategy, the proposed design leads to more distinct feature maps for rich data augmentation, leading to an effective solution to the small sample problem. In addition, due to utilization of the DNN-based cascade structure, more abstract and robust semantics are explored from large scale data sets. As a result, the proposed ODMTCNet model yields a high quality feature representation on image data sets of different scales. This key finding will be experimentally validated in the next section.
\begin{table*}
\normalsize
     \begin{center}
      \caption{ Experimental Settings and Results on the ORL Database}
        \begin{tabular}{|c|c|c|c|c|c|c|c|c|}
        \hline
         \bf Methods & \bf Patch Size & \bf Layers & \bf Filters & \bf \emph{A} Blocks & \bf{$\eta$} & \bf 120 (Training) & \bf 200 (Training) & \bf 280 (Training)
         \\
        \hline
        \hline

 & \bf {$3\times 3$}  &  2 & $L_1$ = $L_2$ = $8$ & 8 & 0.5 & 83.89 $\pm$ 4.72 & 94.55 $\pm$ 2.10 & 97.83 $\pm$ 1.63\\
     CCANet  & \bf {$3\times 3$}  &  3 & $L_1$ = $L_2$ = $L_3$= $8$ & 8 & 0.5 & 86.64 $\pm$ 3.89 & 94.70 $\pm$ 1.97 & 97.92 $\pm$ 0.90\\
\hline
\hline
 & \bf {$3\times 3$}  &  2 & $L_1$ = $L_2$ = $8$ & 8 & 0.5 & 85.96 $\pm$ 0.04 & 95.45 $\pm$ 2.03 & 98.25 $\pm$ 1.07\\
  ODMTCNet & \bf {$3\times 3$}  &  3 & $L_1$ = $L_2$ = $L_3$= $8$ & 8 & 0.5 & \bf{87.89 $\pm$ 3.90} & \bf{95.80 $\pm$ 1.83} & \bf{98.50 $\pm$ 1.23}\\

        \hline
        \end{tabular}
        \end{center}
        \end{table*}
\section{Experimental Results and Analysis}
To validate the effectiveness of ODMTCNet, we conduct experiments on two different image-based tasks: face recognition and object recognition. To demonstrate the generality of the proposed network model, experiments are conducted on four data sets of different scale levels: small scale level, medium scale level and comparatively large scale level. Specifically, experiments are implemented on the Olivetti Research Lab (ORL) database (small level) [87], Facial Recognition Technology (FERET) database (small level) [88] for face recognition; the ETH--80 database (medium level) [89] and Caltech 256 (comparatively large level) database [90] for object recognition. In this work, we conducted experiments with different numbers of convolutional filters, and the best performance is reported. The `Training \#' in all tables denotes the number of training samples.
\subsection{The ORL database}
In this subsection, to examine the performance of ODMTCNet in face recognition, especially on a small scale database, we conduct experiments on the ORL database. In the ORL database, there are 40 subjects with 10 different samples for each person. The samples are captured under different conditions of illumination, facial expression, posing, etc. Some examples from the ORL are shown in Figure. 9.
\begin{figure}[H]
\centering
\includegraphics[height=1.5in,width=3.0in]{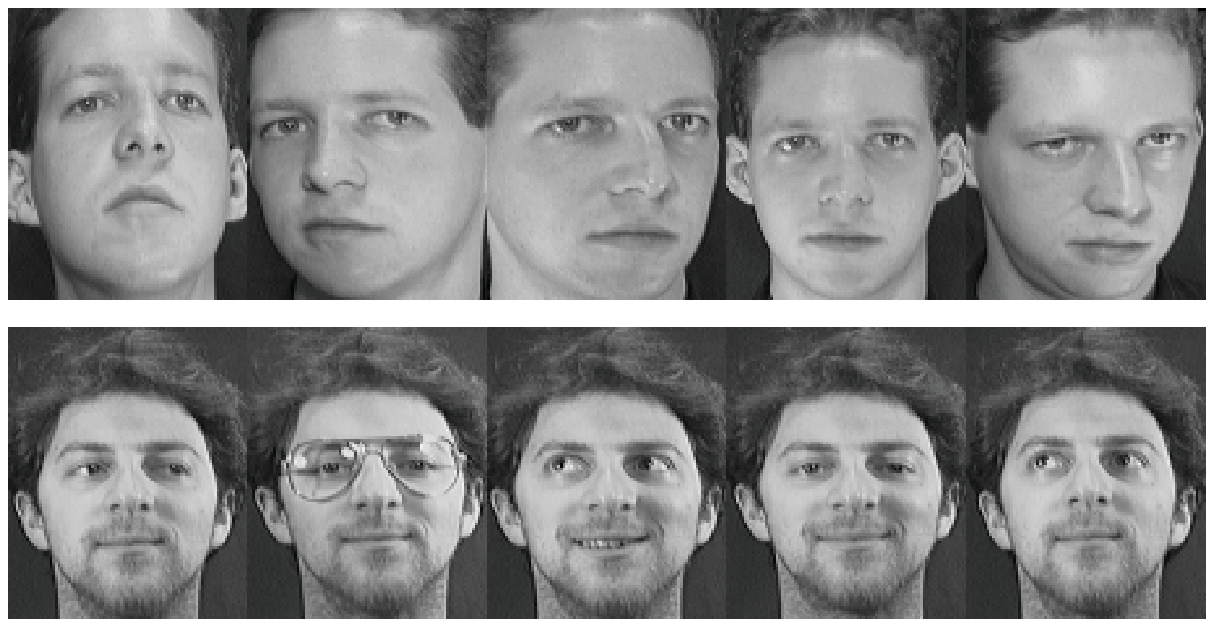}\\ Figure. 9 Example images in the ORL database\\
\end{figure}
During our experiments, the local binary patterns (LBP) [17] operation is performed on each sample to generate the second view data set for ODMTCNet. As a result, the original images and LBP maps are utilized jointly to accomplish the task of multi-view feature representation for face recognition as shown in Figure. 10.\\ \hspace*{\fill} \\
\centerline {\includegraphics[height=2.1in,width=3.6in]{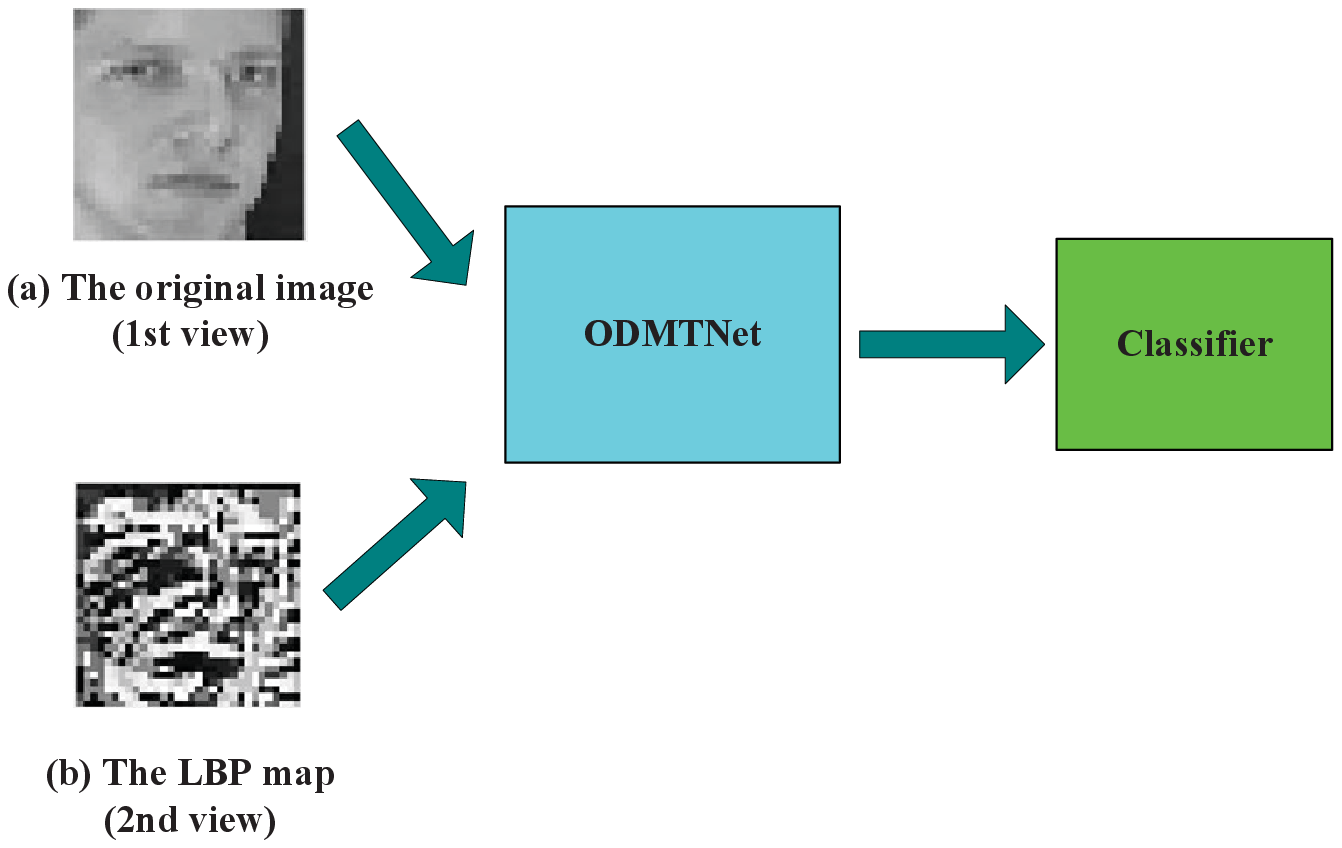}}\\ \centerline {Figure 10. The diagram of ODMTCNet on the ORL database}\\

All 400 samples in the ORL database are utilized, including the training subset and testing subset, respectively. In our experiments, we randomly select 120, 200 and 280 images to be training samples, while the remaining images are used as the testing samples. We run all experiments 10 times. 
To demonstrate the effectiveness of the ODMTCNet, the CCANet is also performed on the ORL database. The experimental settings and results (Average $\pm$ Standard Deviation) are tabulated in TABLE 1. According to TABLE 1, it is observed that ODMTCNet always achieves better performance than that of CCANet with the same experimental settings. In addition, the average accuracy of ODMTCNet is improved with the number of training samples increasing. The comparison amongst state-of-the-art algorithms [8, 11, 18-33] is presented in TABLE 2, showing that ODMTCNet outperforms both statistical machine learning (SML) and DNN based algorithms. Moreover, it is seen that the number of all three filters $L_i$ (i=1, 2, 3) corresponding to the best average performance equals 8 (e.g., $L_1=L_2=L_3=8$) as shown in TABLE 1. This is substantially less than 40, the number of classes. \\
\vspace*{-10pt}
\begin{table}[h]
\normalsize
\renewcommand{\arraystretch}{1.3}
\caption{\normalsize{Comparison with other algorithms on the ORL database}}
\setlength{\abovecaptionskip}{0pt}
\setlength{\belowcaptionskip}{10pt}
\centering
\tabcolsep 0.07in
\begin{tabular}{ccc}
\hline
\hline
Methods & Training \#  & Accuracy\\
\hline
\hline
MCPCADP [18] & 200 & 91.25\%\\
MDL [19] & 200 &92.15\%\\
WLCRC [20] & 200 & 92.69\%\\
GLS [21] & 200 & 93.33\%\\
MSDL [22] & 200 &92.15\%\\
GDLMPP[23] & 200 &94.50\%\\
CCA [24] & 200 &94.50\% \\
MCCA [25] & 200 &94.50\% \\
DSR [26] & 200  &95.00\% \\
SOLDE-TR [27] & 200 &95.03\%\\
CDPL [28] & 200 &95.42\%\\
\bf{ ODMTCNet}($L_i$$<$$40$) & 200 & \bf{95.80\%}\\
\hline
\hline
ANFIS-ABC [29] & 280 &96.00\%\\
ESPP [30] & 280 & 96.00\%\\
CDM [31] & 280 & 95.80\%\\
CNN [8] & 280 &95.92\%\\
PCANet [11] & 280 &96.28\%\\
IKLDA+PNN [32]  & 280 &96.35\%\\
LiSSA [33] & 280 & 97.51\%\\
\bf{ ODMTCNet} ($L_i$$<$$40$) & 280 & \bf{98.50\%}\\
\hline
\hline
\end{tabular}
\end{table}
\subsection{The FERET database}
\begin{table*}
     \begin{center}
      \caption{ Experimental Settings and Results on the FERET Database}
        \begin{tabular}{|c|c|c|c|c|c|c|c|c|}
        \hline
         \bf Methods & \bf Patch Size & \bf Layers & \bf Filters & \bf \emph{A} Blocks & \bf {$\eta$} & \bf Front-Left & \bf Front-Right
         \\
        \hline
        \hline

 & \bf {$3\times 3$}  &  2 & $L_1$ = $L_2$ = $8$ & 4 & 0.5 & 93.50 & 90.00 \\
     CCANet & \bf {$3\times 3$}  &  3 & $L_1$ =14, $L_2$=$L_3$=$7$ & 8 & 0.5 & 95.50 & 91.50 \\
\hline
\hline
 & \bf {$3\times 3$}  &  2 & $L_1$ = $L_2$ = $8$ & 4 & 0.5 & 96.50 & 92.00 \\
 ODMTCNet & \bf {$3\times 3$}  &  3 & $L_1$ =14, $L_2$=$L_3$=$7$ & 8 & 0.5 & \bf{96.50} & \bf{93.00}\\

        \hline
        \end{tabular}
        \end{center}
        \end{table*}
In the FERET database, 600 samples of 200 subjects are chosen. Each subject provides three samples with a size of 20 $\times$ 20 pixels according to three different poses (front, left and right) as shown in Figure. 11.
\begin{figure}[H]
\centering
\includegraphics[height=1.8in,width=1.5in]{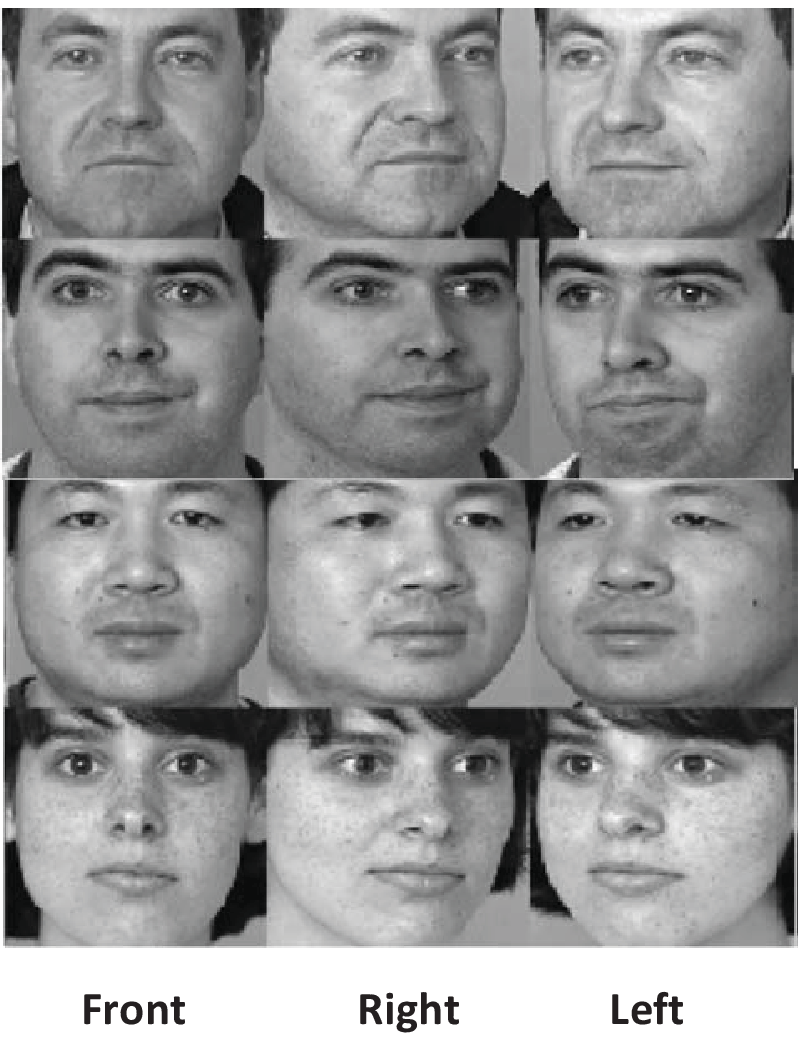}\\ Figure. 11 Examples with different poses from the FERET database\\
\end{figure}
Next, these samples are utilized to construct the first view data set and the twice wavelet transform [34] is performed on each sample in the first data set to generate the corresponding second data set. Moreover, two experimental settings are adopted, front-left and front-right. In the first one, the front samples are utilized for training and the others for testing. In the second, the front images are still adopted as train samples but the right samples are utilized for testing. Essentially, in this subsection, our task is to apply ODMTCNet to two-view single sample based face recognition, which is a challenging and popular research topic. Experimental settings and results on CCANet and ODMTCNet are tabulated in TABLE 3.\\\indent Furthermore, other two-view based methods [35-39] are applied to the above samples to verify the effectiveness of the proposed method. Since the two dimensional principal component analysis (2DPCA) [35] and two dimensional linear discriminant analysis (2DLDA) [36] algorithms are not able to explore the correlation between the two variable sets directly, they are applied to the first view data set only. Consequently, two dimensional CCA (2DCCA) [37], local two dimensional CCA (L2DCCA) [38] and complete discriminative tensor representation learning (CDTRL) [39] are performed on the two-view 2D data sets directly while samples are reshaped into one dimensional vectors for CCA [24]. Then, the recognition results are reported in TABLE 4.\\
\vspace*{-10pt}
\begin{table}[h]
\normalsize
\renewcommand{\arraystretch}{1.4}
\caption{\normalsize{Recognition results with other algorithms on the FERET dataset}}
\setlength{\abovecaptionskip}{0pt}
\setlength{\belowcaptionskip}{10pt}
\centering
\tabcolsep 0.07in
\begin{tabular}{ccc}
\hline
\hline
Method & Front-Left & Front-Right \\
\hline
\hline
2DPCA [35]  &78.50\% &76.50\% \\
2DLDA [36] &71.50\% &70.50\% \\
CCA [24] &72.50\% &68.50\% \\
2DCCA--FFS1 [37]  &80.50\% &74.50\%\\
2DCCA--FFS2 [37]  &80.50\% &74.00\%\\
L2DCCA [38]  &79.50\% &75.00\%\\
CDTRL [39]  &83.00\% &78.00\%\\
\bf{ODMTCNet} ($L_i$$<$$200$) &\textbf{96.50\%} &\textbf{93.00\%}\\
\hline
\hline
\end{tabular}
\end{table}\\\indent
According to TABLE 4, it is observed that ODMTCNet outperforms the other two-view feature representation based methods, leading to greatly improved performance with a large margin. It validates the generality and effectiveness of the proposed ODMTCNet on the small scale data sets. Moreover, it is clearly seen in TABLE 3 that the number of filters associated with the optimal result equals 14 and 7 (e.g., $L_1=14, L_2=L_3=7$). Again, this is far less than 200, the number of classes.
\begin{table*}
     \begin{center}
      \caption{ Experimental Settings and Results on the ETH-80 Database}
      \scalebox{0.95}{
        \begin{tabular}{|c|c|c|c|c|c|c|c|c|c|}
        \hline
         \bf Methods & \bf Patch Size & \bf Layers & \bf Filters & \bf \emph{A} Blocks & \bf {$\eta$} & \bf 500 (Training) & \bf 1000 (Training) & \bf 1500 (Training) & \bf 1640 (Training)
         \\
        \hline
        \hline

 & \bf {$7\times 7$}  &  2 & $L_1$ = $L_2$ = $8$ & 7 & 0.5 & 86.05 $\pm$ 0.89 & 91.97 $\pm$ 0.61 & 93.62 $\pm$ 0.61 & 93.98 $\pm$ 0.21\\
     CCANet  & \bf {$7\times 7$}  &  3 & $L_1$ = $L_2$ = $8$, $L_3$= $4$ & 7 & 0.5 & 87.20 $\pm$ 0.76 & 91.51 $\pm$ 0.47 & 93.51 $\pm$ 0.51 & 93.17 $\pm$ 0.28\\
\hline
\hline
 & \bf {$7\times 7$}  &  2 & $L_1$ = $L_2$ = $8$ & 7 & 0.5 & \bf{88.28 $\pm$ 0.49} & \bf{92.24 $\pm$ 0.60} & \bf{94.19 $\pm$ 0.61} & \bf{94.40 $\pm$ 0.20}\\
 ODMTCNet & \bf {$7\times 7$}  &  3 & $L_1$ = $L_2$ = $8$, $L_3$= $4$ & 7 & 0.5 & 88.16 $\pm$ 0.91 & 92.19 $\pm$ 0.79 & 93.60 $\pm$ 0.27 & 93.69 $\pm$ 0.12\\

        \hline
        \end{tabular}}
        \end{center}
        \end{table*}
\subsection{The ETH--80 database}
As a popular data set for multi-view feature representation studies, the ETH--80 database consists of 3280 color RGB object samples. All samples are divided into eight classes, including apples, cars, cows, cups, dogs, horses, pears and tomatoes. For each class, there are 410 images captured for 10 different objects and each object is presented by 41 images from a variety of angles. Some examples are shown in Figure. 12.
\begin{figure}[H]
\centering
\includegraphics[height=1.5in,width=3.3in]{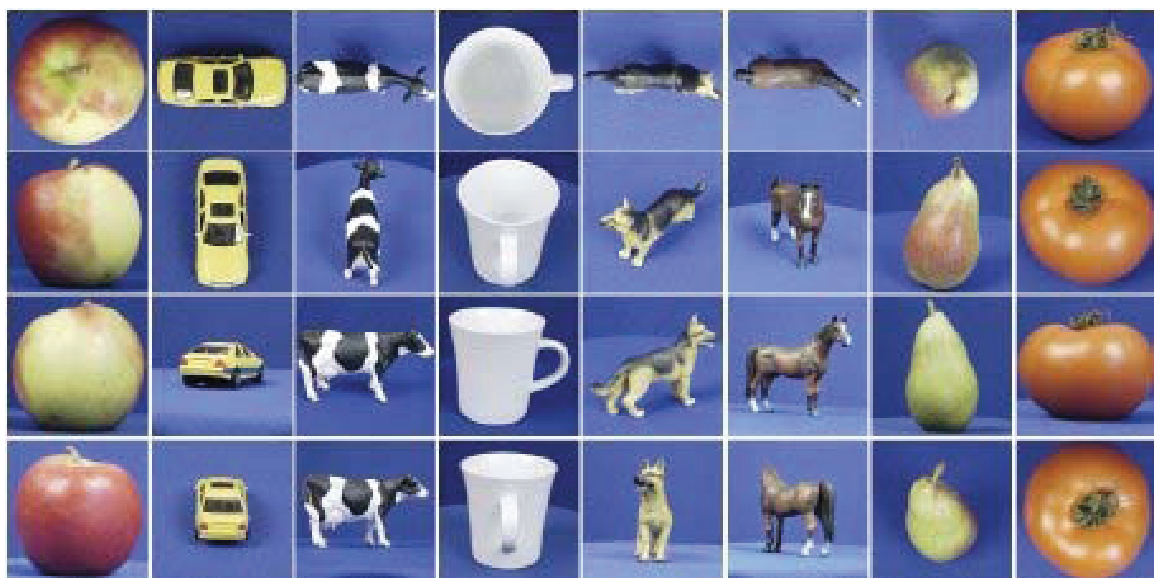}\\ Figure. 12 Some examples from the ETH--80 database\\
\end{figure}
In our experiments, all samples are normalized to the size of 64 $\times$ 64 pixels. The R and G sub-channel images are adopted as the two different views, as shown in Figure. 13. Moreover, we randomly choose 500, 1000, 1500 and 1640 images to construct the training subset while the remaining images are used as the testing subset. All experiments are conducted 10 times and the average results are reported. To further verify the effectiveness of the ODMTCNet, the CCANet is also applied to the ETH-80 database. The experimental settings and results (Average $\pm$ Standard Deviation) are tabulated in TABLE 5.
\begin{figure}[H]
\centering
\includegraphics[height=1.1in,width=3.6in]{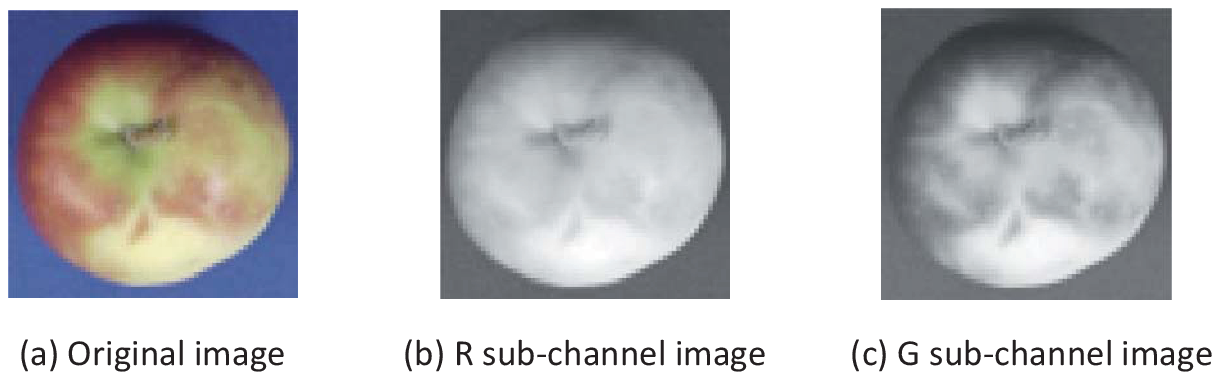}\\ Figure. 13 The original image, R and G sub-channel images in the ETH--80 database\\
\end{figure}
For comparison, recognition rates with state-of-the-art methods are listed in TABLE 6. From TABLE 5 and TABLE 6, it is seen that ODMTCNet outperforms CCANet under the same parameter settings. In addition, ODMTCNet achieves better performance than that of state-of-the-art algorithms. Furthermore, it is observed that ODMTCNet obtains higher recognition accuracy than the algorithms in [40, 41, 42] even with less training samples, validating its effectiveness on image feature representation. Again, the number of filters $L_i$ (i=1, 2) associated with the optimal performance is 8 (e.g., $L_1 = L_2 = 8$) as given in TABLE 5, less than the number of classes 80.\\
\vspace*{-10pt}
\begin{table}[h]
\normalsize
\renewcommand{\arraystretch}{1.2}
\caption{\normalsize{Comparison with other state-of-the-art on the ETH--80 database}}
\setlength{\abovecaptionskip}{0pt}
\setlength{\belowcaptionskip}{10pt}
\centering
\tabcolsep 0.07in
\begin{tabular}{ccc}
\hline
\hline
Methods & Training \#  & Accuracy\\
\hline
\hline
LapMCC [40] & 984 & 69.50\%\\
HesMCC [41] & 984 &73.52\%\\
\bf{ODMTNet}($L_i$$<$$80$) & 500 & \bf{88.18\%}\\
\hline
\hline
ALP-TMR [42] & 1400 &88.92\% \\
\bf{ODMTNet}($L_i$$<$$80$) & 1000 & \bf{92.24\%}\\
\hline
\hline
Fine-tuned AlexNet [43] & 1640 & 94.20\%\\
SDNN [44] & 1640 &82.80\%\\
MMFML-M3 [45] & 1640 & 92.53\%\\
MKDR-LR [46] & 1640 &91.20\%\\
Kernel-LP [47] & 1600 &89.11\% \\
CMCM [48] & 1640  &92.50\% \\
JDRML [49] & 1640 &94.00\%\\
MCCM-LE [50] & 1640 &93.30\%\\
SML [51] & 1640 &94.02\%\\
Sparsity+Intra-task [52] & 1640 & 92.50\%\\
TLRDA+PCA [53] & 1640 & 92.00\%\\
SSL-TR [54] & 1640 &93.40\%\\
PLSRGStO [55]  & 1640 &92.50\%\\
SRC+DPC [56] & 1640 & 94.00\%\\
RMML-GM [57] & 1640 & 90.25\%\\
LRRTDR [58] & 1640 & 92.00\%\\
rBDLR [59] & 2624 & 93.55\%\\
PDOD+AOD+KNN [60] & 3239 & 93.25\%\\
\bf{ODMTNet}($L_i$$<$$80$) & 1640 & \bf{94.40\%}\\
\hline
\hline
\end{tabular}
\end{table}
\subsection{The Caltech 256 database}
\begin{table}[h]
\normalsize
\renewcommand{\arraystretch}{1.2}
\caption{\normalsize{Performance of \textbf{fc6} and \textbf{fc7} on the Caltech256 database}}
\setlength{\abovecaptionskip}{0pt}
\setlength{\belowcaptionskip}{10pt}
\centering
\tabcolsep 0.07in
\begin{tabular}{cc}
\hline
\hline
Feature & Recognition Accuracy\\
\hline
\hline
\textbf{fc6} & 58.22\% \\
\textbf{fc7} & 59.81\% \\
\hline
\hline
\end{tabular}
\end{table}
\begin{table*}
     \begin{center}
      \caption{ Experimental Settings and Results on the Caltech 256 Database}
        \begin{tabular}{|c|c|c|c|c|c|c|c|c|c|}
        \hline
         \bf Methods & \bf Patch Size & \bf Layers & \bf Filters & \bf \emph{A} Blocks & \bf {$\eta$} & \bf 30/class$\times$257=7710 (Training) & \bf 60/class$\times$257=15420 (Training)
         \\
        \hline
        \hline

 & \bf {$5\times 5$}  &  2 & $L_1$ = $L_2$ = $8$ & 8 & 0.5 & 84.81 $\pm$ 1.00 & 87.20 $\pm$ 0.40\\
     CCANet  & \bf {$5\times 5$}  &  3 & $L_1$ = $L_2$ = $8$, $L_3$= $4$ & 8 & 0.5 & 85.07 $\pm$ 0.44 & 87.82$\pm$ 0.33\\
\hline
\hline
 & \bf {$5\times 5$}  &  2 & $L_1$ = $L_2$ = $8$ & 8 & 0.5 & 86.27 $\pm$ 0.31 & 87.25 $\pm$ 0.31\\
 ODMTCNet & \bf {$5\times 5$}  &  3 & $L_1$ = $L_2$ = $8$, $L_3$= $4$ & 8 & 0.5 & \bf{86.33 $\pm$ 0.26} & \bf{88.34 $\pm$ 0.35}\\

        \hline
        \end{tabular}
        \end{center}
        \end{table*}
\begin{figure}[H]
\centering
\includegraphics[height=1.3in,width=3.2in]{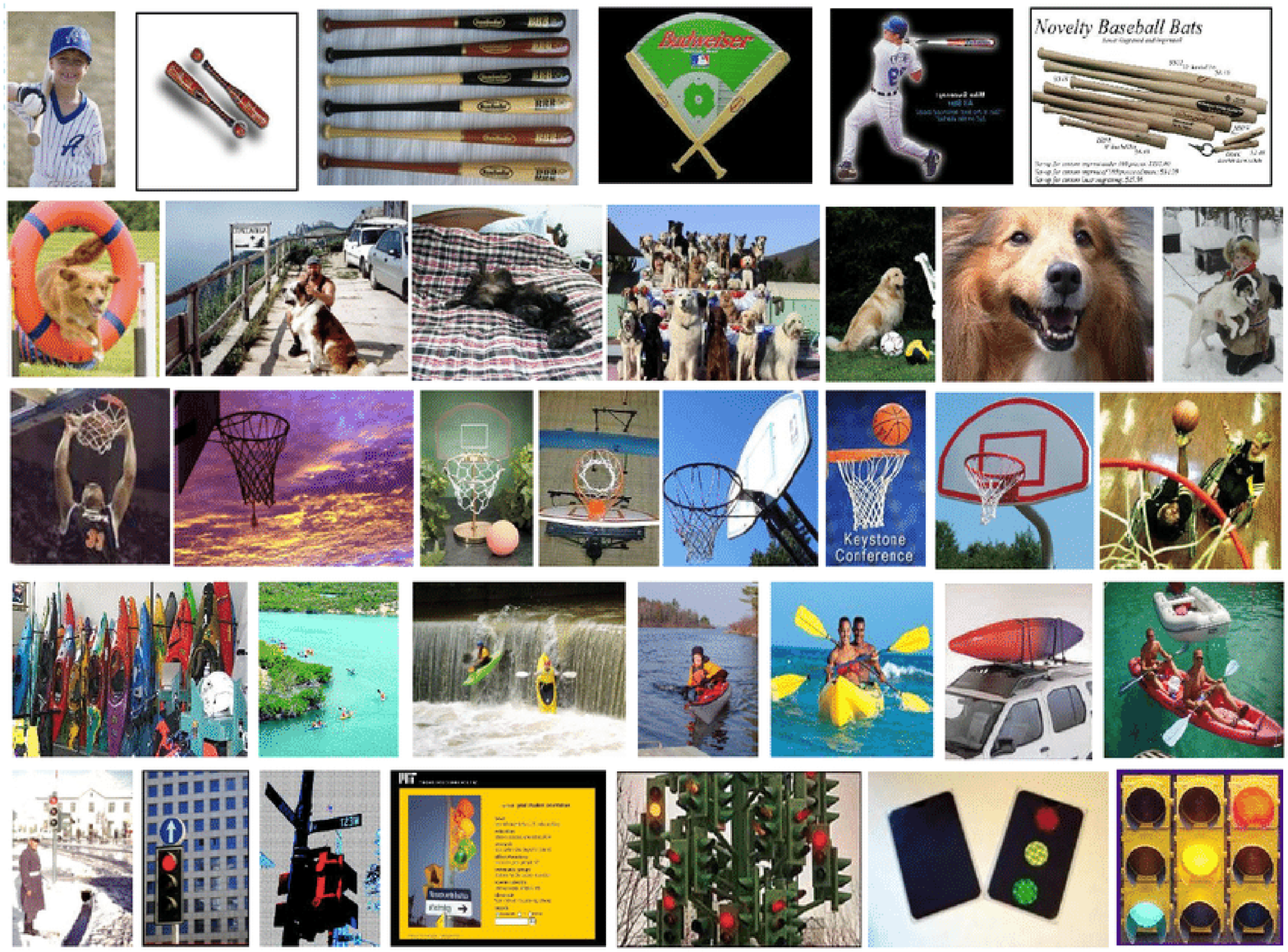}\\ Figure. 14 Some samples from Caltech 256 database
\end{figure}
Caltech 256 database contains a varying set of illumination, movements, backgrounds, etc. The classes are hand-picked to represent a wide variety of natural and artificial objects in various settings. Several sample images from the Caltech 256 are given in Figure. 14.\\
\vspace*{-10pt}
\begin{table}[h]
\normalsize
\renewcommand{\arraystretch}{1.2}
\caption{\normalsize{Recognition accuracy with other methods on the Caltech 256}}
\setlength{\abovecaptionskip}{0pt}
\setlength{\belowcaptionskip}{10pt}
\centering
\tabcolsep 0.07in
\begin{tabular}{ccc}
\hline
\hline
Methods & Training \# & Accuracy \\
\hline
\hline
CMFA-SR [62] &7710 &71.44\% \\
DGFLP [63] &7710 &69.17\% \\
BMDDL [65] &7710 &59.30\% \\
LLKc [66] &7710 &72.09\% \\
ISC-LG [67] &7710 &50.62\% \\
BLF-FV [68] &7710 &51.42\% \\
OCB-FV [69] &7710 &53.15\% \\
LLC-SVM [70] &7710 &34.50\% \\
SWSS-VGG [71] &7710 &73.56\% \\
ResNet152 [72] &7710 &78.00\% \\
ResFeats152 [73]  &7710 &79.50\% \\
MVLS [74] &7710 &84.23\% \\
NR [75] &7710 &84.40\% \\
TransTailor [64] &7710 &85.30\% \\
Hybrid1365-VGG [76] &7710 &76.04\% \\
LMCCA [77] &7710 &76.52\% \\
PtR [78] &7710 &84.50\% \\
DMCCA [15]  &7710 &80.32\% \\
Autuencoder-DenseNet-121 [95] &7710 &80.60\% \\
HMML [79]  &12850 &64.06\% \\
DCS [80] &13770 &64.18\% \\
\textbf{ODMTCNet ($L_i$$<$$257$}) &7710 & \textbf{86.33\%} \\
\hline
\hline
CMFA-SR [62] &15420 &76.31\% \\
SMNN+VGG-19 [81] &15420 &81.90\% \\
SMNN+Resnet152 [81] &15420 &82.30\% \\
SMNN+Xception [81] &15420 &84.70\% \\
LLKc [66] &15420 &75.36\% \\
ISC-LG [67] &15420 &55.76\% \\
OCB-FV [69] &15420 &59.03\% \\
LLC-SVM [70] &15420 &40.10\% \\
SWSS-VGG [71] &15420 &76.25\% \\
ResNet152 [72] &15420 &81.90\% \\
ResFeats152 [73]  &15420 &82.10\% \\
NAC [82] &15420 &84.10\% \\
Joint fine-tuning [83] &15420 &83.80\% \\
DFG [80] &15420 &87.09\% \\
\textbf{ODMTCNet ($L_i$$<$$257$}) &15420 & \textbf{88.34\%} \\
\hline
\hline
\end{tabular}
\end{table}\\\indent
For fair comparison, we set the same experimental settings as in previous studies. Specifically, 30 images and 60 images are chosen from each class as training samples. To verify the generic nature and the power of the proposed solution, a relatively simple DNN model, VGG--19 [61], is employed to extract DNN based features. Two fully connected layers \textbf{fc6} and \textbf{fc7} are used for extracting the multi-view features. Since the dimensional number of \textbf{fc6} and \textbf{fc7} is 4096, both of them are reshaped as two-dimensional maps in the size of $64 \times 64$, with the performance tabulated in TABLE 7. Afterwards, the two-dimensional maps from \textbf{fc6} and \textbf{fc7} are utilized as two-view data sets for CCANet and ODMTCNet. All experiments are conducted 10 times and the results (Average $\pm$ Standard Deviation) are reported. Experimental settings and results are given in TABLE 8. Moreover, the performance of state-of-the-art methods [15, 62-83, 95] is reported in TABLE 9. Again, the generated feature representation outperforms other methods in object recognition by a significant margin, especially when more than 15000 samples are used in training. Note, the number of filters $L_i$ (i=1, 2, 3) corresponding to the best performance is 8 and 4 (e.g., $L_1 = L_2 = 8, L_3 = 4$) shown in TABLE 8. This is much less than the number of classes 257 in the Caltech 256 database.
\subsection{Discussions}
The presented mathematical analysis and experimental validation clearly demonstrate that the proposed ODMTCNet operates well on image data sets of different scale levels, outperforming all the comparing algorithms as shown in Tables 2, 4, 6 and 9. Furthermore, the following desirable properties of ODMTCNet are observed:\\\indent

a) The network is able to take image features in all popular formats as input: pixel values in ORL and ETH-80, hand-crafted features in FERET and DNN features in Caltech 256.\\\indent
b) The number of filters ($L_i$) leading to the best accuracy is much lower than the number of classes, validating the desirable relation presented in Section 3.4. In fact, the number of convolutional parameters corresponding to the filters is on the scale of several hundreds (as shown in Tables 1, 3, 5, and 8), stunningly different form that of mainstream DNNs (on the scale of tens of millions up to several billions).\\

With drastically smaller number of convolutional parameters, fit for training on CPU scale processors, the studies reported in this paper potentially open up a new and more accessible path for the promotion of interpretable machine learning (I-ML) in both academic research and practical applications.

\section{Conclusions}
In this paper, an optimal discriminant multi-view tensor convolutional network (ODMTCNet) is proposed for deep level multi-view feature representation. By integrating the SML principles with DNN architecture, the parameters of convolutional layers are determined by an analytical solution instead of the traditional BP algorithm. Hence, the proposed ODMTCNet forms a foundational platform for an interpretable machine learning model which generates feature representation of high quality from original multi-view data sources. Furthermore, the representation effectively addresses the overfitting problem when there is no sufficient data available in training. The desirable properties of ODMTCNet are validated by numerous application examples with image data sets of different scales. To conclude this paper, it is worth noting that the number of convolutional layers corresponding to the optimal performance in ODMTCNet is smaller than or equals to 3 in all the four evaluation examples, approaching the bound predicted in Kolmogorov Superposition Theorem [93-94] in neural network-based representation.


%





\ifCLASSOPTIONcaptionsoff
  \newpage
\fi

\end{document}